\documentclass{bmvc2k}

\usepackage{booktabs}
\usepackage{multirow}

\title{\fontsize{13}{16}\selectfont Mitigating Bias with Words: Inducing Demographic Ambiguity in Face Recognition Templates by Text Encoding}

\addauthor{Tahar Chettaoui}{tahar.chettaoui@igd.fraunhofer.de}{1}
\addauthor{Naser Damer}{naser.damer@igd.fraunhofer.de}{1,2}
\addauthor{Fadi Boutros}{fadi.boutros@igd.fraunhofer.de}{1}

\addinstitution{
Fraunhofer IGD\\
Fraunhoferstraße 5\\
64283 Darmstadt
}
\addinstitution{
TU Darmstadt \\
Karolinenpl. 5,\\
64289 Darmstadt
}

\runninghead{Tahar Chettaoui, Naser Damer, Fadi Boutros}{Mitigating Bias with Words}

\begin{document}

\maketitle
\setlength{\textfloatsep}{8pt}
\vspace{-8pt}

\begin{abstract}
Face recognition (FR) systems are often prone to demographic biases, partially due to the entanglement of demographic-specific information with identity-relevant features in facial embeddings. This bias is extremely critical in large multicultural cities, especially where biometrics play a major role in smart city infrastructure. The entanglement can cause demographic attributes to overshadow identity cues in the embedding space, resulting in disparities in verification performance across different demographic groups. To address this issue, we propose a novel strategy, Unified Text-Image Embedding (UTIE), which aims to induce demographic ambiguity in face embeddings by enriching them with information related to other demographic groups. This encourages face embeddings to emphasize identity-relevant features and thus promotes fairer verification performance across groups. UTIE leverages the zero-shot capabilities and cross-modal semantic alignment of Vision-Language Models (VLMs). Given that VLMs are naturally trained to align visual and textual representations, we enrich the facial embeddings of each demographic group with text-derived demographic features extracted from other demographic groups. This encourages a more neutral representation in terms of demographic attributes. We evaluate UTIE using three VLMs, CLIP, OpenCLIP, and SigLIP, on two widely used benchmarks, RFW and BFW, designed to assess bias in FR. Experimental results show that UTIE consistently reduces bias metrics while maintaining, or even improving in several cases, the face verification accuracy.
\end{abstract}

\vspace{-12pt}
\section{Introduction} \label{sec:intro}
\vspace{-8pt}
Face recognition (FR) systems are increasingly integrated into real-world applications \cite{DBLP:journals/ijon/WangD21a, jones2021law, DBLP:journals/corr/abs-2505-24247}, yet concerns around demographic bias remain a barrier to their fair and reliable deployment \cite{DBLP:journals/corr/abs-2003-02488, DBLP:journals/tbbis/JainDE22, DBLP:journals/tasm/RathgebDFDB22, DBLP:conf/aies/RajiGMBLD20}. Such biases manifest as disparities in recognition accuracy across demographic groups, including race, gender, and age \cite{jones2021law, DBLP:journals/corr/abs-2502-02309, DBLP:conf/fat/BuolamwiniG18, DBLP:conf/cvpr/SVKAB19, DBLP:conf/wacv/AlbieroSVZKB20}. Addressing these disparities is essential to ensure equitable FR performance, motivating the development of methods \cite{DBLP:conf/eccv/AlviZN18, DBLP:journals/corr/abs-1911-10692, DBLP:journals/corr/abs-2004-11246, PAMI_BIAS, DBLP:conf/iwbf/TerhorstTDKK20} to mitigate bias while maintaining high recognition accuracy. Despite progress in developing bias mitigation techniques, effectively reducing disparities across demographic groups in verification scenarios remains challenging. This challenge arises partly because embeddings used for face verification may encode demographic-specific attributes that contribute to biased performance across groups \cite{DBLP:conf/iccv/DharGRCC21, nist_250171}. In this work, we address this issue, focusing on disparities in verification performance across demographic groups.

\begin{figure}
\begin{center}
\begin{tabular}{cc}
\bmvaHangBox{\fbox{\includegraphics[width=4.5cm]{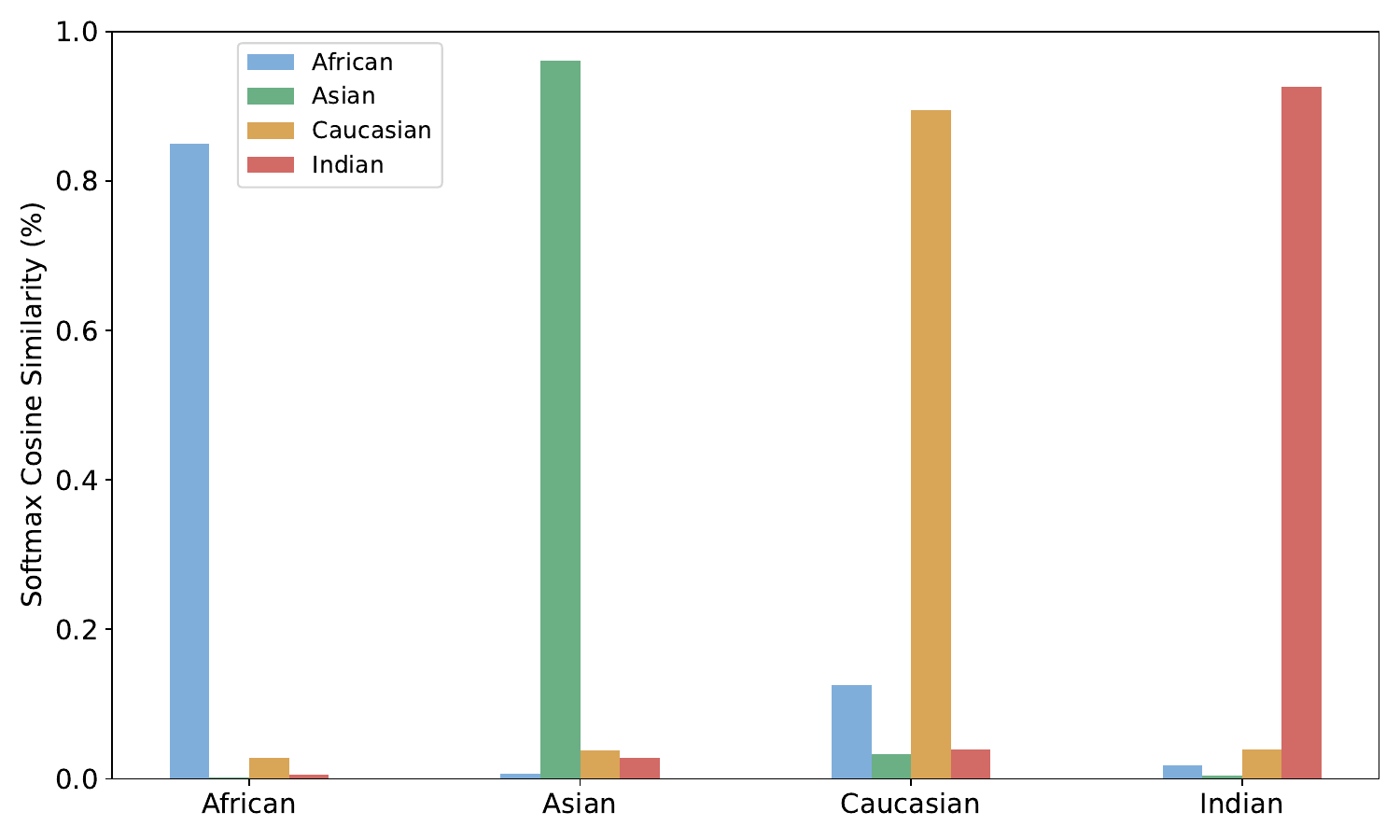}}}&
\bmvaHangBox{\fbox{\includegraphics[width=4.5cm]{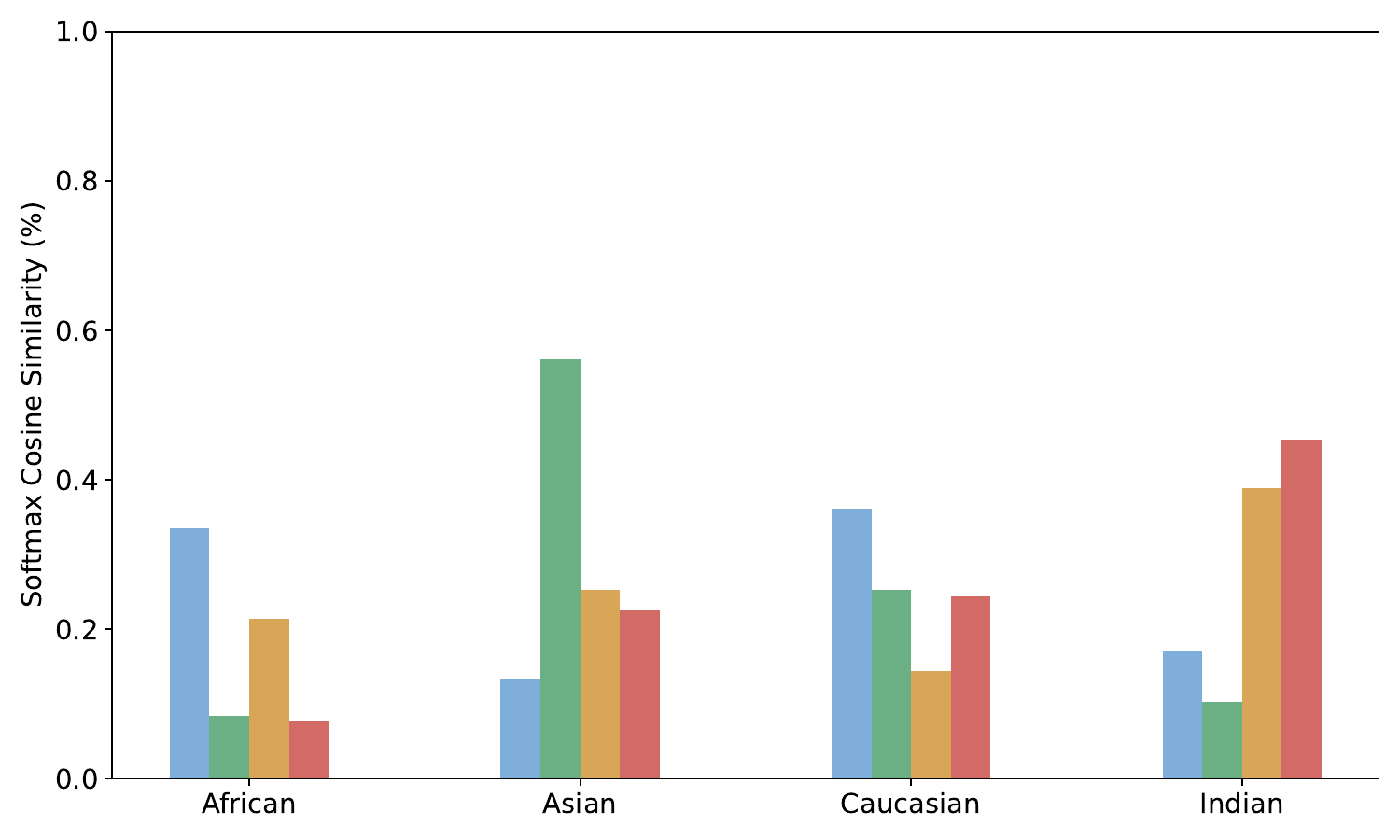}}}\\
(a) Baseline (IE), with the template defined as \( I \) & (b) UTIE, with the template defined as \( I' \)
\end{tabular}
\end{center}
\caption{Comparison of our UTIE average cosine similarities with racial text embeddings against the baseline (IE) on the RFW subsets, using CLIP ViT/B-16. For each subset, we compute the average cosine similarities of the face embeddings with the different text embeddings \( T_{i} \), for \(i \in {1,..,4}  \), representing the four considered races: African (blue), Asian (green), Caucasian (yellow), and Indian (red). (a) IE displays a high similarity with the corresponding demographic class \( T_{\hat{i}} \) for all subsets while maintaining lower similarities with the other classes \( T_{i} \)  for \(i \in {1,..,4}, \; i \neq \hat{i}  \). This shows that face embeddings encode demographic-specific information. (b) For UTIE, the similarity to \( T_{\hat{i}} \) is significantly lower compared to IE, and the similarities are spread more evenly across the other classes. This indicates that UTIE makes the racial information in the embeddings more ambiguous, reducing the alignment with a single demographic and distributing similarity more evenly across classes.}
\label{fig:cosine_sim}
\end{figure}

Since verification relies on comparing embeddings extracted from face images, we hypothesize that one contributing factor to this bias is the presence of demographic-specific information in these embeddings. To mitigate this, we propose a novel approach that encourages face embeddings to be more demographically ambiguous by enriching them with features representative of all demographic groups. 
This strategy aims to make the learned representations more generalizable and less sensitive to demographic attributes, thereby reducing bias in face verification performance. To achieve this, we leverage multimodal Vision-Language Models (VLMs) and the fact that they produce aligned embeddings for image and text. We use the text encoder to obtain embeddings that capture demographic characteristics. Next, we exclude the embedding corresponding to the predicted characteristic (e.g., racial, gender), i.e, the textual embedding that matches the image embedding with the highest similarity score. We then compute the average of the remaining embeddings and add this mean vector to the original image embedding. By incorporating features representative of other demographic groups into the final visual feature representation, we aim to make the influence of demographic characteristics more ambiguous within the face embedding. The evaluations include demographic bias analysis on the Racial Faces in the Wild (RFW) \cite{DBLP:conf/iccv/WangDHTH19} and Balanced Faces In the Wild (BFW) \cite{DBLP:conf/cvpr/RobinsonLHQ0T20} datasets, reporting verification accuracies along with bias assessment metrics, including the Skewed Error Ratio (SER) and standard deviation (STD). Both datasets evaluate racial bias, and BFW additionally includes gender bias evaluation. Furthermore, we evaluate the demographic prediction performance in a zero-shot setting on the same datasets since our approach relies on accurate and consistent demographic attribute predictions from the text encoder to construct the enriched face embeddings. We considered three VLMs, namely CLIP \cite{DBLP:conf/icml/RadfordKHRGASAM21}, OpenCLIP \cite{DBLP:conf/cvpr/ChertiBWWIGSSJ23}, and SigLIP \cite{DBLP:conf/iccv/ZhaiM0B23}, and our results reveal that our unified text-image embedding (UTIE) successfully reduces the STD and SER while improving or maintaining comparable mean accuracy across all models on both racial and gender bias evaluation. For instance, in the racial bias evaluation, baseline CLIP achieves a mean accuracy of 72.20\% on RFW, which increases to 72.25\% using UTIE, while the STD decreases using our UTIE from 4.81 to 4.46 and the SER from 1.50 to 1.45. Similarly, in the gender bias evaluation on BFW, using UTIE with CLIP reduces the STD from 2.72 to 2.58 and lowers the SER from 1.28 to 1.27, while maintaining comparable accuracy. These results confirm that UTIE enhances demographic ambiguity in the embeddings, leading to reduced bias while preserving or improving recognition performance. These findings contribute to ongoing research on bias mitigation in FR and support the development of demographically ambiguous face embeddings as a viable strategy for reducing bias. 

\vspace{-12pt}
\section{Related Work} \label{sec:rw}
\vspace{-8pt}
Demographic bias in FR has received significant attention. Several works \cite{DBLP:journals/corr/abs-2502-02309, DBLP:conf/wacv/AlbieroSVZKB20, DBLP:journals/tifs/KlareBKBJ12} examined the causes of demographic bias, identifying imbalanced or unrepresentative datasets as a factor. Albiero et al. \cite{DBLP:conf/wacv/AlbieroSVZKB20} identified the under-representation of females in training data as a key factor contributing to gender bias. Another study \cite{DBLP:journals/tifs/KlareBKBJ12} showed that training FR systems on balanced datasets is critical for reducing vulnerabilities on specific demographic groups. Another factor is the variability in skin tone \cite{DBLP:conf/fat/BuolamwiniG18, DBLP:journals/tbbis/LuCCC19}. Buolamwini and Gebru \cite{DBLP:conf/fat/BuolamwiniG18} evaluated three gender classification systems and identified under-performance in recognizing dark-skinned females. A quantitative assessment across six skin tone groups found light-skinned individuals easiest to verify and darker-skinned individuals the most challenging \cite{DBLP:journals/tbbis/LuCCC19}. Other factors, like the impact of image attributes associated with socially constructed gender norms, have also been investigated. Prior works \cite{DBLP:conf/btas/DantchevaCR12, DBLP:journals/tcsv/GuoWY14, ueda_makeup} explored the impact of makeup on FR accuracy, finding that variations in makeup between images make it more challenging to recognize genuine pairs. Other studies \cite{DBLP:conf/cvpr/WuBBB23, DBLP:conf/wacv/WuTBORB24} examined facial hair, showing that similar beard areas increase similarity scores, while differences reduce them. 

All these papers motivated research into how to mitigate bias by modifying the training strategy or the network structure. To promote fairness in FR, \cite{DBLP:journals/corr/abs-1911-10692} proposes a reinforcement learning-based Race Balance Network (RL-RBN) that uses adaptive margins to balance FR performance across races. By modeling margin selection for non-Caucasians as a Markov decision process and applying deep Q-learning, RL-RBN reduces feature distribution skewness. Alvi et al. \cite{DBLP:conf/eccv/AlviZN18} propose an algorithm that removes multiple sources of variation from the feature representation of a network to reduce bias and improve classification accuracy. They introduce a supervised learning approach that learns features informative for the primary task while remaining uninformative for spurious variations such as pose or gender, thereby reducing bias in the learned embeddings. Dhar et al. \cite{DBLP:conf/iccv/DharGRCC21} proposed Protected Attribute Suppression System (PASS), an adversarial debiasing approach to train a model to classify identities while discouraging it from predicting protected attributes. It can be trained on descriptors from any trained high-performing network to classify identities while reducing the encoding of sensitive attributes, effectively mitigating gender and skin tone bias. Another work \cite{PAMI_BIAS} proposed a meta-learning algorithm, Meta Balanced Network (MBN), to mitigate algorithmic bias, which learns adaptive margins in large margin loss. MBN balances the feature scatter among skin tone groups in feature space rather than merely adjusting loss weights.

None of the previous works addressed mitigating bias in VLMs, especially those that are not fine-tuned to the task, which we address in this work. Additionally, and most importantly, this paper is the first to theorize reducing demographic bias in FR by making the face embeddings more demographically ambiguous, and doing that by inducing embeddings encoded from text to the face image embedding. Unlike prior works that encourage balance during training \cite{DBLP:journals/corr/abs-1911-10692, PAMI_BIAS}, discourage the encoding of demographic attributes in the feature representation \cite{DBLP:conf/eccv/AlviZN18, DBLP:conf/iccv/DharGRCC21}, or manipulate the comparison and decision-making processes \cite{DBLP:journals/prl/TerhorstKDKK20,DBLP:conf/iwbf/TerhorstTDKK20}, our approach enriches each face embedding with information derived from embeddings that explicitly capture characteristics of other demographic groups. This balances the influence of demographic attributes and reduces their dominance within the embeddings, resulting in a more demographically ambiguous face representation and thus reducing demographic bias.

\vspace{-12pt}
\section{Methodology} \label{sec:methodology}
\vspace{-8pt}

This section introduces our approach to reducing demographic bias in VLM-based FR by making face embeddings more demographically ambiguous. We first present baseline VLMs, which learn to align images and text embeddings, allowing us to leverage complementary semantic information from text. We then introduce UTIE, showing how we fuse image and text embeddings to obscure demographic information, aiming at reducing demographic bias.

\vspace{-12pt}
\subsection{Baseline VLMs (preliminary)} \label{baseline_vlms}
\vspace{-8pt}

\begin{figure*}
\begin{center}
\includegraphics[width=.8\linewidth]{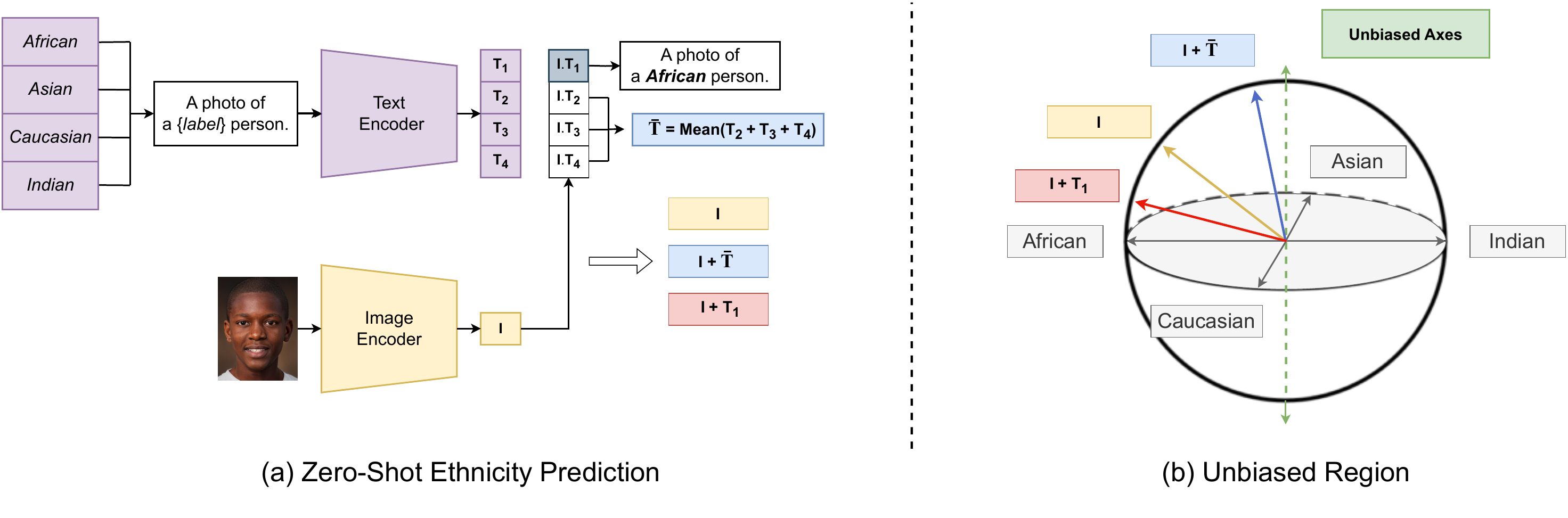}
\end{center}
\vspace{-14pt}
\caption{Demographically ambiguous embedding generation with UTIE for racial bias mitigation. (a) We use a text encoder \( f_t(\cdot) \) to extract racial embeddings \( T_{i} \), for \(i \in {1,..,4}  \) for zero-shot prediction. We then exclude the predicted embedding \( T_{\hat{i}} \) (in this case \( T_{1} \)), compute the average of the remaining demographic embeddings, denoted as \( \bar{T} \), and add it to the original image embedding \( I \) resulting to \( I' \). (b) The yellow vector represents the baseline IE. 
The blue vector represents our UTIE. 
This vector is closer to the unbiased axes, as it is more demographically ambiguous, as demonstrated in Section \ref{sec:unified_text_image}. Finally, the red vector represents IE+PTE 
defined in Section \ref{sec:experimentalsetup}. This vector shows stronger demographic dominance, in this case leaning more toward African identity, illustrating increased demographic influence.}
\label{fig:bias_concept}
\end{figure*}

In this paper, we considered three VLMs, namely CLIP \cite{DBLP:conf/icml/RadfordKHRGASAM21}, OpenCLIP \cite{DBLP:conf/cvpr/ChertiBWWIGSSJ23}, and SigLIP \cite{DBLP:conf/iccv/ZhaiM0B23}. Some of which have been shown to be usable as template extractors for FR \cite{DBLP:journals/ivc/ChettaouiDB25}. These methods are based on contrastive learning, which aligns related image-text pairs by minimizing the distance between their visual and textual embeddings in a shared space, while maximizing the distance between embeddings of unrelated pairs \cite{DBLP:conf/icml/RadfordKHRGASAM21, DBLP:conf/cvpr/ChertiBWWIGSSJ23, DBLP:conf/icml/JiaYXCPPLSLD21, DBLP:conf/icml/0001LXH22}. Given a batch of $N$ (image, text) pairs, the model is trained to identify which of the $N \times N$ possible pairings correspond to the true pairs. To achieve this, they jointly train an image encoder \( f_i(\cdot) \) and a text encoder \( f_t(\cdot) \) to align the image and text embeddings for $N$ matching (positive) image-text pairs while making sure that unrelated $N_2 - N$ (negative) image-text pairs are dissimilar in the embedding space. This is done by maximizing the cosine similarity of the image and text embeddings of the matching pairs in the batch while minimizing the cosine similarity of the embeddings of the incorrect pairings.
SigLIP is trained using a sigmoid loss instead of the softmax-based contrastive loss. It operates solely on individual image-text pairs and does not require a global view of pairwise similarities for normalization. This allows for scaling to larger batch sizes and improves performance at smaller batch sizes. The nature of these pre-trained VLMs allows the extraction of embeddings from face images that can be used for face verification \cite{DBLP:conf/iccv/ZhaiM0B23}. They also enable generating embeddings from text that are aligned with those of the face images. This alignment between text and image embeddings forms the basis of our UTIE solution, as described in the next Section \ref{sec:unified_text_image}.

\vspace{-12pt}
\subsection{Unified Text-Image Embedding (UTIE)} \label{sec:unified_text_image}
\vspace{-8pt}
This section presents our UTIE, which combines image and text embeddings to obscure demographic-specific information and reduce demographic bias. Since face verification compares embeddings that may encode demographic cues, we propose an approach that encourages face embeddings to be more demographically ambiguous by enriching them with features representative \( T_{i} \) of all \( N \) demographic groups. Specifically, we enrich each face embedding with information derived from embeddings that are explicitly designed to capture characteristics of other (not dominant in this face) demographic groups. To achieve this, we leverage VLMs, which align image and text embeddings in the same space, allowing us to enrich face embeddings with demographic features, as detailed in Section \ref{baseline_vlms}.

As depicted in Figure \ref{fig:bias_concept}.a, given a sample face image \( x \), we pass it through the image encoder to obtain its image embedding \( I \). Next, we aim to generate features that are representative of all demographic groups. To achieve this, we build on the alignment within the image-text embedding space and propose leveraging the text encoder to generate these representative features. Following the zero-shot settings described in Section \ref{sec:experimentalsetup}, we extract text embeddings, denoted as \( T_{i} \), for the \( N \) class labels corresponding to different demographic characteristics. Since the prompt specifies only the demographic classes, the text embeddings \( T_{i} \), for \(i \in {1, .., N}  \), capture information that is ideally exclusively related to those demographics. We then use these embeddings for zero-shot prediction, where we compute the predicted demographic class as the one with the maximum cosine similarity to \( I \):

\vspace{-10pt}
\begin{equation}
\hat{i} = \arg\max_{i} \frac{\langle I, T_{i} \rangle}{\| I \| \cdot \| T_{i} \|},
\label{eq:ihat}
\end{equation}
\vspace{-14pt}

As a result, we obtain the index \( \hat{i} \), corresponding to the predicted demographic class. This index indicates the demographic group that the model predicts for \( I \) based on the maximum alignment in the embedding space. To encourage demographic ambiguity, we aim to shift the image embedding away from this dominant demographic group. To do that, we need to enrich the face image embedding with information about the other classes by constructing the mean text embedding across all demographic groups, excluding the predicted one:

\vspace{-10pt}
\begin{equation}
\bar{T} = \frac{1}{N-1} \sum_{i \neq \hat{i}} T_{i},
\label{eq:bart}
\end{equation}
\vspace{-14pt}

where \( N \) is the number of demographic classes. We then add \( \bar{T} \) to the image embedding to obtain the final demographically ambiguous representation using our UTIE:

\vspace{-10pt}
\begin{equation}
I' = I + \bar{T}.
\label{eq:Istrich}
\end{equation}
\vspace{-20pt}

In our UTIE solution, we propose to use this adjusted embedding \( I'\) as the face template for verification instead of the original embedding \( I \). By adding the mean embedding \( \bar{T} \) to the image embedding \( I \), this operation aims to reduce the dominance of the predicted demographic class \( T_{\hat{i}} \). As a result, \( I' \) distributes more evenly across the directions associated with all demographic groups. This makes the resulting feature vector \( I' \) less aligned with any single demographic group and more balanced across all groups, making it demographically ambiguous and helping to reduce demographic bias in downstream tasks.

Figure \ref{fig:cosine_sim} compares the cosine similarity distributions of image embeddings with racial text embeddings across the four demographic classes on the RFW \cite{DBLP:conf/iccv/WangDHTH19}, introduced in the next section, subsets for both the baseline (IE) and our proposed approach (UTIE). For all subsets, IE displays a high similarity with the corresponding demographic class \( T_{\hat{i}} \) while maintaining notably lower similarities with the other classes \( T_{i} \)  for \(i \in {1, .., N}, \; i \neq \hat{i}  \). On the other hand, for our approach (UTIE), the similarity to \( T_{\hat{i}} \) is significantly lower compared to IE, and the similarities spread more evenly across other classes. This indicates that our method makes the racial information in the embeddings more ambiguous, reducing the strong alignment with a single demographic and distributing similarity more evenly across classes.

\vspace{-12pt}
\section{Experimental Setup} \label{sec:experimentalsetup}
\vspace{-8pt}
\textbf{Face Template Setups.} To validate our approach, we consider three feature representations. The first configuration uses the unmodified image embedding (IE) as a baseline, with the template defined as \( I \) in Section \ref{sec:unified_text_image}. The second configuration applies our proposed UTIE approach, with the template defined as \( I' \) in Section \ref{sec:unified_text_image}. This aims to mitigate bias, making demographic influences more ambiguous in the representation. As a third configuration, to demonstrate that performing the counter concept to UTIE increases bias, we introduce the IE+PTE setup, where PTE stands for Predicted demographic Text Embedding. In this setup, instead of aiming at reducing demographic information, we attempt to reinforce it by adding the predicted demographic text embedding \( T_{\hat{i}} \) to the image embedding \( I \), resulting in:

\vspace{-10pt}
\begin{equation}
I^{*} = I + T_{\hat{i}}.
\label{eq:Istar}
\end{equation}
\vspace{-19pt}

This configuration allows us to examine whether explicitly incorporating the predicted demographic attribute amplifies demographic bias in the embeddings, providing a contrasting perspective that highlights the effectiveness of our UTIE approach in reducing bias.

\textbf{Evaluation Datasets.}
We evaluate the considered VLMs on the RFW \cite{DBLP:conf/iccv/WangDHTH19} and BFW \cite{DBLP:conf/cvpr/RobinsonLHQ0T20} datasets to assess model bias and performance across demographic groups, as they are widely used for fairness and bias evaluation \cite{DBLP:journals/ivc/ChettaouiDB25, PAMI_BIAS}. The RFW dataset contains four testing subsets corresponding to African, Asian, Caucasian, and Indian groups.
The BFW dataset extends this evaluation by providing eight subgroups balanced across gender and race, enabling a more detailed assessment of intersectional bias. For racial bias evaluation, both RFW and BFW are suitable, while gender bias is assessed only with BFW.

\textbf{Zero-Shot Settings.} Following the evaluation benchmarks \cite{DBLP:conf/iccv/WangDHTH19, DBLP:conf/cvpr/RobinsonLHQ0T20} for racial bias evaluation, we consider four ethnicities as classes: African, Asian, Caucasian, and Indian. CLIP \cite{DBLP:conf/icml/RadfordKHRGASAM21}, OpenCLIP \cite{DBLP:conf/cvpr/ChertiBWWIGSSJ23}, and SigLIP \cite{DBLP:conf/iccv/ZhaiM0B23} were trained using sentence-based prompts rather than single-word labels. This motivates using prompts during zero-shot evaluation, such as \texttt{"A photo of a \{label\}."}, to help bridge the distribution gap between pre-training and downstream tasks, which has been shown to improve accuracy. Additionally, it was observed that zero-shot performance can be significantly enhanced by customizing the prompt text for each task \cite{DBLP:conf/icml/RadfordKHRGASAM21}. For these reasons, we adopt the prompt \texttt{"A photo of a \{label\} person."} as our standard prompt in this work. Following the same setup, we evaluate gender bias in Section \ref{sec:gender_bias_eval} using two classes, female and male, instead of four ethnic groups.

\textbf{Evaluation Metrics.}
We follow the reporting protocols and evaluation metrics associated with the evaluation datasets and recent works \cite{DBLP:journals/ivc/ChettaouiDB25, PAMI_BIAS, DBLP:conf/iccv/WangDHTH19}. We report the results as verification accuracies in (\%) on each subset and as average accuracies to evaluate general recognition performance on the benchmarks. To evaluate the bias, we report the STD between all subsets and the SER, which is given by $\frac{max_gError_g}{min_gError_g}$, where $g$ represents the demographic group, as reported in \cite{PAMI_BIAS, DBLP:journals/corr/abs-1911-10692}. A higher STD value indicates more bias across demographic groups and vice versa. For SER, models with values closer to 1 are less biased.

\textbf{Model Architectures.} We use official pre-trained models for all three VLMs: CLIP (ViT-B/16) \cite{DBLP:conf/icml/RadfordKHRGASAM21}, OpenCLIP (ViT-B/16) \cite{DBLP:conf/cvpr/ChertiBWWIGSSJ23}, and SigLIP (ViT-B/16) \cite{DBLP:conf/iccv/ZhaiM0B23}. The OpenCLIP model was trained on the LAION-2B dataset \cite{DBLP:conf/nips/SchuhmannBVGWCC22}, which includes a 2.32 billion-sample English image-text subset. For comparison, CLIP \cite{DBLP:conf/icml/RadfordKHRGASAM21} was trained on a smaller dataset of 400 million (image, text) pairs sourced from large-scale publicly available data. For SigLIP,  we use the ViT-B/16 model trained on the WebLI dataset \cite{DBLP:conf/iclr/Chen0CPPSGGMB0P23} at a resolution of 224×224.


\aboverulesep=0ex
\belowrulesep=0ex

\vspace{-12pt}
\section{Results} \label{sec:results}
\vspace{-8pt}

This section presents the results achieved by our UTIE solution on the considered VLMs on RFW and BFW. We evaluate three feature representation strategies, IE, UTIE, and IE+PTE, presented in Section \ref{sec:experimentalsetup}. We assess these configurations for racial bias in Section \ref{sec:racialbias} and gender bias in Section \ref{sec:gender_bias_eval}, examining their impact on fairness in verification performance.


\vspace{-12pt}
\subsection{Racial Bias in FR Evaluation} \label{sec:racialbias}
\vspace{-8pt}
We evaluated the considered approaches, namely CLIP \cite{DBLP:conf/icml/RadfordKHRGASAM21}, OpenCLIP \cite{DBLP:conf/cvpr/ChertiBWWIGSSJ23}, and SigLIP \cite{DBLP:conf/iccv/ZhaiM0B23}, on the RFW \cite{DBLP:conf/iccv/WangDHTH19} and BFW \cite{DBLP:conf/cvpr/RobinsonLHQ0T20} benchmarks to assess model bias and performance across different demographic groups. To evaluate our proposed UTIE, we compare it with the baseline (IE) and a validation setup (IE+PTE) where we add the predicted demographic embedding to amplify bias signals. This setup allows us to systematically evaluate the impact of each feature representation on racial bias. One can observe the following from the results in Table \ref{tab:racial_bias}, highlighting the impact of different feature representations on racial bias:

\textbf{Comparison of UTIE vs. IE:} UTIE reduces STD and SER across all settings on both datasets, while improving mean accuracy on RFW \cite{DBLP:conf/iccv/WangDHTH19} and maintaining comparable accuracy on BFW \cite{DBLP:conf/cvpr/RobinsonLHQ0T20}. For instance, CLIP \cite{DBLP:conf/icml/RadfordKHRGASAM21} achieves a mean accuracy of 72.20\% on RFW with IE and 72.25\% with UTIE, while reducing STD from 4.81 to 4.46 and SER from 1.50 to 1.45. On BFW, CLIP with UTIE maintains a high mean accuracy (84.23\% vs. 84.50\%) while reducing STD (1.54 vs. 1.63) and SER (1.26 vs. 1.29). This trend holds for OpenCLIP \cite{DBLP:conf/cvpr/ChertiBWWIGSSJ23} and SigLIP \cite{DBLP:conf/iccv/ZhaiM0B23}, demonstrating that UTIE, by making face embeddings more demographically ambiguous as shown in Figure \ref{fig:cosine_sim}, can preserve performance and reduce bias. In contrast, the IE+PTE setup yields higher STD and SER, confirming that adding predicted demographic embeddings amplifies demographic bias as intended for validation.

\textbf{Comparison across models:} CLIP \cite{DBLP:conf/icml/RadfordKHRGASAM21} and OpenCLIP \cite{DBLP:conf/cvpr/ChertiBWWIGSSJ23} consistently achieve higher average accuracy across both RFW \cite{DBLP:conf/iccv/WangDHTH19} and BFW \cite{DBLP:conf/cvpr/RobinsonLHQ0T20} compared to SigLIP \cite{DBLP:conf/iccv/ZhaiM0B23}, with BFW means around 84\% for both CLIP and OpenCLIP, while SigLIP reaches around 79\%. SigLIP exhibits lower STD and SER on BFW, reaching an STD of 0.97 with UTIE compared to 1.54 and 2.13 for CLIP and OpenCLIP, respectively, indicating stronger consistency across demographic groups despite its lower overall accuracy. Compared to OpenCLIP, CLIP is more balanced across demographic groups, exhibiting lower STD and SER in all cases.

\textbf{Comparison of Dataset influence on accuracy and bias:} Across all models, the average accuracy is consistently higher on BFW \cite{DBLP:conf/cvpr/RobinsonLHQ0T20} than on RFW \cite{DBLP:conf/iccv/WangDHTH19}, consistent with previous works \cite{DBLP:conf/cvpr/XuHSLLHL021, DBLP:conf/wacv/Fournier-Montgieux25, DBLP:conf/wacv/HuberLBKD24}. For example, baseline CLIP \cite{DBLP:conf/icml/RadfordKHRGASAM21} achieves an average accuracy of 84.50\% on BFW compared to 72.20\% on RFW. Additionally, both the STD and SER are lower on BFW across all models, indicating reduced demographic variance compared to RFW.

Overall, our UTIE consistently reduced the bias of the investigated model over racial groups, while having minimal effect on the FR performance. This, along with the clear negative effect of the counter concept, IE+PTE, validates our approach of reducing demographic bias by making the demographic information in the embedding more ambiguous. 

\begin{table}[ht]
\label{tab:racial_bias}
\begin{center}
\scriptsize 
\resizebox{\textwidth}{!}{
\begin{tabular}{cc|
cccc|ccc|
cccc|ccc}
\toprule
\multirow{2}{*}{\textbf{Approach}} & \multirow{2}{*}{\textbf{Feature Embedding}} 
& \multicolumn{7}{c|}{\textbf{RFW}\cite{DBLP:conf/iccv/WangDHTH19}} 
& \multicolumn{7}{c}{\textbf{BFW}\cite{DBLP:conf/cvpr/RobinsonLHQ0T20}} \\
\cmidrule(lr){3-9} 
\cmidrule(lr){10-16} 
& & African & Asian & Caucasian & Indian & Mean & STD & SER & Asian & Black & Indian & White & Mean & STD & SER  \\ \midrule

\multirow{3}{*}{CLIP\cite{DBLP:conf/icml/RadfordKHRGASAM21}} 
    & IE & 70.75 & 69.73 & 79.32 & 68.98 & 72.20 & 4.81 & 1.50 & 82.36 & 84.49 & 84.85 & 86.31 & \textit{84.50} & 1.63 & 1.29 \\ 
    & UTIE & 70.85 & 69.80 & 78.88 & 69.48 & \textit{72.25} & \textbf{4.46} & \textbf{1.45} & 82.20 & 84.16 & 84.68 & 85.89 & 84.23 & \textbf{1.54} & \textbf{1.26} \\ 
    & IE+PTE  & 68.47 & 68.73 & 77.90 & 66.87 & 70.49 & 5.01 & 1.50 & 82.22 & 83.37 & 83.96 & 86.25 & 83.95 & 1.70 & 1.29\\ \hline

\multirow{3}{*}{OpenCLIP\cite{DBLP:conf/cvpr/ChertiBWWIGSSJ23}} 
    & IE &  69.37 & 68.60 & 79.95 & 69.72 & 71.91 & 5.38 & 1.57 & 80.49 & 86.01 & 83.97 & 86.35 & \textit{84.20} & 2.69 & 1.43\\
    & UTIE &  69.35 & 68.83 & 79.80 & 69.85 & \textit{71.96} & \textbf{5.24} & \textbf{1.54} & 80.54 & 85.23 & 83.79 & 84.85 & 83.60 & \textbf{2.13} & \textbf{1.32} \\ 
    & IE+PTE & 67.83 & 67.62 & 78.93 & 65.58 & 69.99 & 6.05 & 1.63 & 80.07 & 84.85 & 82.63 & 86.37 & 83.48 & 2.74 & 1.46 \\ \hline

\multirow{3}{*}{SigLIP\cite{DBLP:conf/iccv/ZhaiM0B23}} 
    & IE &  58.17 & 64.17 & 71.62 & 65.98 & 64.98 & 5.54 & 1.47 &  78.27 & 79.52 & 80.57 & 80.54 & \textit{79.73} & 1.09 & 1.12 \\
    & UTIE & 58.63 & 64.62 & 71.63 & 65.60 & \textit{65.12} & \textbf{5.32} & 1.46 &  77.83 & 78.91 & 79.93 & 79.80 & 79.12 & \textbf{0.97} & \textbf{1.10} \\
    & IE+PTE & 56.80 & 61.02 & 69.86 & 63.53 & 62.80 & 5.46 & \textbf{1.43} &  77.77 & 79.17 & 79.93 & 80.88 & 79.44 & 1.31 & 1.16 \\

\bottomrule
\end{tabular}
}
\end{center}
\caption{Evaluation results of different feature representations for the considered models on RFW \cite{DBLP:conf/iccv/WangDHTH19} and BFW \cite{DBLP:conf/cvpr/RobinsonLHQ0T20} reported as average verification accuracy in (\%), STD, and SER across four different demographic groups. The higher mean indicates better verification accuracy, and the higher STD indicates a more biased model. For SER, the model that achieved a value closer to 1 is less biased. UTIE consistently reduces STD and SER across all settings on both datasets, while maintaining comparable accuracy on RFW and BFW.}
\end{table}

\vspace{-12pt}
\subsection{Gender Bias Evaluation} \label{sec:gender_bias_eval}
\vspace{-8pt}

Besides racial bias, we assess gender bias using the BFW dataset \cite{DBLP:conf/cvpr/RobinsonLHQ0T20}. We follow the same evaluation settings used for the racial bias assessment and examine the same approaches, focusing on their performance across the two gender groups. From Table \ref{tab:gender_bias}, highlighting the impact of different feature representations on gender bias, the following can be noted:


\textbf{Comparison of UTIE vs. IE:} For all models, UTIE shows only a slight decrease in average recognition performance compared to baseline IE. Additionally, UTIE consistently reduces both STD and SER across the two gender groups on BFW, demonstrating that increasing demographic ambiguity in embeddings can effectively reduce gender bias. For example, with UTIE, OpenCLIP \cite{DBLP:conf/cvpr/ChertiBWWIGSSJ23} improves, with STD dropping from 3.86 to 3.06 and SER from 1.42 to 1.30. CLIP \cite{DBLP:conf/icml/RadfordKHRGASAM21} STD decreases from 2.72 to 2.58, with a slight reduction in SER from 1.28 to 1.27. SigLIP \cite{DBLP:conf/iccv/ZhaiM0B23} maintains a stable STD while showing a small SER improvement. The IE+PTE configuration consistently demonstrates lower accuracy as well as higher STD and SER, confirming that incorporating predicted demographic text embeddings amplifies demographic influence and bias in representations, as expected for validation. These observations follow the trend observed in the racial bias evaluation in Section \ref{sec:racialbias}.

\textbf{Comparison across models:} CLIP \cite{DBLP:conf/icml/RadfordKHRGASAM21} generally achieves the highest mean accuracy, followed by OpenCLIP \cite{DBLP:conf/cvpr/ChertiBWWIGSSJ23}, with SigLIP \cite{DBLP:conf/iccv/ZhaiM0B23} showing lower performance. OpenCLIP reports a higher STD of 3.06 and SER of 1.30 than CLIP, which achieves an STD of 2.58 and SER of 1.27, indicating that OpenCLIP is slightly more biased across genders, although UTIE helps reduce this gap. SigLIP consistently yields lower mean accuracy but demonstrates the lowest STD of 1.59 and SER of 1.11, reflecting lower gender bias.

Our UTIE reduces gender bias in the evaluated model without significantly impacting FR accuracy, similar to the trend observed in the racial bias evaluation in Section \ref{sec:racialbias}. 

\begin{table}[ht]
\label{tab:gender_bias}
\begin{center}
\tiny
\begin{tabular}{cc|cc|ccc}
\toprule
\textbf{Approach} & \textbf{Feature Embedding} & Female & Male & Mean & STD & SER \\
\midrule
\multirow{3}{*}{CLIP\cite{DBLP:conf/icml/RadfordKHRGASAM21}} 
    & IE & 82.58 & 86.43 & 84.50 & 2.72 & 1.28 \\ 
    & UTIE & 82.58 & 86.23 & 84.41 & \textbf{2.58} & \textbf{1.27} \\ 
    & IE+PTE & 82.65 & 86.58 & \textit{84.61} & 2.78 & 1.29 \\  \hline 

\multirow{3}{*}{OpenCLIP\cite{DBLP:conf/cvpr/ChertiBWWIGSSJ23}} 
    & IE & 81.48 & 86.93 & \textit{84.20 }& 3.86 & 1.42 \\ 
    & UTIE & 81.44 & 85.76 & 83.60 & \textbf{3.06} & \textbf{1.30} \\ 
    & IE+PTE & 81.25 & 86.74 & 83.99 & 3.88 & 1.41 \\ \hline 
    
\multirow{3}{*}{SigLIP\cite{DBLP:conf/iccv/ZhaiM0B23}} 
    & IE & 78.60 & 80.85 & \textit{79.73 }& 1.59 & 1.12 \\ 
    & UTIE & 78.13 & 80.38 & 79.25 & \textbf{1.59} & \textbf{1.11} \\ 
    & IE+PTE & 78.29 & 80.76 & 79.52 & 1.75 & 1.13 \\ 

\bottomrule
\end{tabular}
\end{center}
\caption{Evaluation results of the different feature representations for the considered models on BFW \cite{DBLP:conf/cvpr/RobinsonLHQ0T20} reported as average verification accuracy in (\%), STD, and SER across the two gender groups. The UTIE consistently reduces STD and SER across all settings.
}
\end{table}

\vspace{-12pt}
\subsection{Zero-shot text encoder performance:} \label{sec:zero-shot-text}
\vspace{-8pt}

Our approach relies on consistent demographic attribute predictions from the text encoder to construct the enriched face embeddings. This is based on the need to determine the demographic group \( \hat{i} \) for the image embedding \( I \), as used in Equation \ref{eq:ihat}. Therefore, following the zero-shot settings detailed in Section \ref{sec:experimentalsetup}, we evaluate the zero-shot prediction accuracy of the text encoder for both race and gender. From Table \ref{tab:textual_bias}, which presents the zero-shot accuracy of the text encoder for racial and gender prediction, we draw the following observations:

\textbf{Analysis of zero-shot text encoder performance in recognizing racial groups:} All models achieve high mean accuracy across groups, with CLIP \cite{DBLP:conf/icml/RadfordKHRGASAM21} and OpenCLIP \cite{DBLP:conf/cvpr/ChertiBWWIGSSJ23} reaching around 92-93\% on RFW \cite{DBLP:conf/iccv/WangDHTH19} and 87.92\% and 91.44\%, respectively, on BFW \cite{DBLP:conf/cvpr/RobinsonLHQ0T20}. In comparison, SigLIP \cite{DBLP:conf/iccv/ZhaiM0B23} shows a lower mean accuracy, scoring 88.07\% on RFW and 86.15\% on BFW, reflecting its overall lower zero-shot text encoder performance. 

\textbf{Analysis of zero-shot text encoder performance in recognizing gender groups:} All models achieve high mean accuracy across gender groups, with CLIP \cite{DBLP:conf/icml/RadfordKHRGASAM21} and OpenCLIP \cite{DBLP:conf/cvpr/ChertiBWWIGSSJ23} reaching 98.42\% and 97.42\% on BFW, respectively, while SigLIP \cite{DBLP:conf/iccv/ZhaiM0B23} shows lower performance at 95.11\%, highlighting its generally lower zero-shot text encoder accuracy. 

\begin{table}[ht]
\label{tab:textual_bias}
\begin{center}
\scriptsize 
\resizebox{\textwidth}{!}{
\begin{tabular}{cc|
cccc|c|
cccc|c|
cc|c}
\toprule
\multirow{2}{*}{\textbf{Approach}} & \multirow{2}{*}{\textbf{Architecture}}
& \multicolumn{5}{c|}{\textbf{RFW}} 
& \multicolumn{8}{c}{\textbf{BFW}} \\
\cmidrule(lr){3-7} 
\cmidrule(lr){8-15} 
& & African & Asian & Caucasian & Indian & Mean & Asian & Black & Indian & White & Mean & Female & Male & Mean\\ \midrule

\multirow{1}{*}{CLIP} 
    & ViT-B/16 & 87.72 & 96.34 & 97.31 & 89.74 & 92.78 & 98.80 & 64.00  & 90.84 & 98.04 & 87.92 & 97.86 & 98.98 & \textbf{98.42}  \\ 
    
\multirow{1}{*}{OpenCLIP} 
    & ViT-B/16 & 95.99 & 93.74 & 97.44 & 84.57 & \textbf{92.94} & 95.16 & 85.40 & 85.88 & 99.32 & \textbf{91.44} &  96.78 & 98.06 & 97.42 \\
    
\multirow{1}{*}{SigLIP} 
    & ViT-B/16 & 92.68 & 82.08 & 96.77 & 80.75 & 88.07 & 83.68 & 79.20 & 82.82 & 98.90 & 86.15 & 92.68 & 97.54 & 95.11 \\ 

\bottomrule
\end{tabular}
}
\end{center}
\caption{Zero-shot racial prediction accuracy on RFW \cite{DBLP:conf/iccv/WangDHTH19} and BFW \cite{DBLP:conf/cvpr/RobinsonLHQ0T20}, and gender prediction accuracy on BFW, for the considered models, reported as average prediction accuracy (\%). OpenCLIP \cite{DBLP:conf/cvpr/ChertiBWWIGSSJ23} achieves the highest mean zero-shot accuracy for racial prediction on RFW and BFW, followed by CLIP \cite{DBLP:conf/icml/RadfordKHRGASAM21}, while SigLIP \cite{DBLP:conf/iccv/ZhaiM0B23} shows the lowest mean accuracy. For gender prediction on BFW, CLIP achieves the highest mean zero-shot accuracy. 
}
\end{table}

Even though performing this recognition is neither our contribution nor our goal, to put the reported results in a wider perspective, the performance compares well to recent works on the same benchmarks, with \cite{DBLP:journals/corr/abs-2412-06235} reporting 93.66\% average face verification accuracy.




\vspace{-12pt}
\section{Conclusion} \label{sec:conclusion}
\vspace{-8pt}
We proposed a novel approach to reduce demographic bias in VLM-based FR by inducing demographic ambiguity in face embeddings. We employed VLMs to create a UTIE that incorporates features from multiple demographic groups to introduce demographic ambiguity, thereby reducing the dominance of any single demographic attribute in the face representation while preserving identity information. We experimentally demonstrated this ambiguity by showing a more evenly distributed demographic similarity distribution on demographic groups after applying our approach, compared to the baseline. Our approach was evaluated on RFW and BFW, using three SOTA models: CLIP, OpenCLIP, and SigLIP. Our results consistently demonstrate a reduction in bias metrics while maintaining verification accuracy. For example, when comparing the baseline with UTIE, we observe that UTIE reduces the STD from 4.81 to 4.46 and the SER from 1.50 to 1.45 on RFW with CLIP while improving the mean accuracy from 72.20\% to 72.25\%. On BFW, CLIP with UTIE maintains a comparable mean accuracy while reducing the STD from 1.63 to 1.54 and the SER from 1.29 to 1.26. This trend holds across OpenCLIP and SigLIP. These findings validate that increasing demographic ambiguity in face embeddings is a promising direction for reducing bias in FR systems and motivate further research on improving this strategy, including prompt engineering, ensembling methods, and investigating its potential impact on privacy leakage.

\vspace{-12pt}
\section*{Acknowledgments}
\vspace{-8pt}
This research work has been funded by the German Federal Ministry of Education and Research and the Hessen State Ministry for Higher Education, Research and the Arts within their joint support of the National Research Center for Applied Cybersecurity ATHENE.

\bibliography{egbib}
\end{document}